\documentclass[10pt,twocolumn,letterpaper]{article}

\usepackage[pagenumbers]{cvpr} % To force page numbers, e.g. for an arXiv version

\usepackage{graphicx}
\usepackage{amsmath}
\usepackage{amssymb}
\usepackage{booktabs}
\usepackage{epsfig}
\usepackage{graphicx}
\usepackage{amsmath}
\usepackage{amssymb}
\usepackage{booktabs}
\usepackage{tabulary}
\usepackage{multirow}
\usepackage{overpic}
\usepackage{xspace}
\usepackage{bbm}
\usepackage{xcolor}         % colors
\usepackage{microtype}
\usepackage{graphicx}
\usepackage{amsmath,amsfonts,bbm}
\usepackage{multirow}
\usepackage{epsfig}
\usepackage{subcaption}
\usepackage{times,graphicx,amsmath,amssymb,booktabs,tabulary,multirow,overpic,xcolor}
\usepackage{mathtools}
\usepackage{wrapfig}
\usepackage{nicefrac}       % compact symbols for 1/2, etc.
\usepackage{microtype}      % microtypography
\usepackage{enumitem}
\usepackage{bbm}

\usepackage[pagebackref=false, breaklinks=true, letterpaper=true, colorlinks,
            citecolor=citecolor, linkcolor=linkcolor, bookmarks=false]{hyperref}
\definecolor{citecolor}{HTML}{0071BC}
\definecolor{linkcolor}{HTML}{ED1C24}

\renewcommand{\paragraph}[1]{\vspace{1.25mm}\noindent\textbf{#1}}

\usepackage[ruled]{algorithm2e}
\usepackage{algorithmic}

\graphicspath{ {./figs/} }
\usepackage{floatrow}
\usepackage{adjustbox}
\newfloatcommand{capbtabbox}{table}[][\FBwidth]

\usepackage[capitalize]{cleveref}
\crefname{section}{Sec.}{Secs.}
\Crefname{section}{Section}{Sections}
\Crefname{table}{Table}{Tables}
\crefname{table}{Tab.}{Tabs.}

\def\cvprPaperID{} % *** Enter the CVPR Paper ID here
\def\confName{CVPR}
\def\confYear{2023}

\usepackage{multirow}

\newcommand{\vct}[1]{\boldsymbol{#1}} % vector
\newcommand{\mat}[1]{\boldsymbol{#1}} % matrix
\newcommand{\methodname}{{TCL}\xspace}
\newcommand{\bd}[1]{\textbf{#1}}
\newcommand{\app}{\raise.17ex\hbox{$\scriptstyle\sim$}}

\newcommand{\std}[1]{{{$\pm$#1}}} % matrix
\newlength\savewidth\newcommand\shline{\noalign{\global\savewidth\arrayrulewidth
  \global\arrayrulewidth 1pt}\hline\noalign{\global\arrayrulewidth\savewidth}}
\newcommand{\T}{^{\textrm T}} % transpose

\begin{document}

%%%%%%%%% TITLE - PLEASE UPDATE
\title{
Twin Contrastive Learning with Noisy Labels
}

% \author{First Author\\
% Institution1\\
% Institution1 address\\
% {\tt\small firstauthor@i1.org}
% % For a paper whose authors are all at the same institution,
% % omit the following lines up until the closing ``}''.
% % Additional authors and addresses can be added with ``\and'',
% % just like the second author.
% % To save space, use either the email address or home page, not both
% \and
% Second Author\\
% Institution2\\
% First line of institution2 address\\
% {\tt\small secondauthor@i2.org}
% }
\author{Zhizhong Huang$^{1}$\qquad Junping Zhang$^{1}$\qquad Hongming Shan$^{2,3}$\thanks{Corresponding author}
\\
$^{1}$ Shanghai Key Lab of Intelligent Information Processing, School of Computer Science,\\
Fudan University, Shanghai 200433, China\\
$^{2}$ Institute of Science and Technology for Brain-inspired Intelligence and MOE Frontiers Center \\for Brain Science,  Fudan University, Shanghai 200433, China\\
$^{3}$ Shanghai Center for Brain Science and Brain-inspired Technology, Shanghai 200031, China\\
{\tt\small \{zzhuang19, jpzhang, hmshan\}@fudan.edu.cn}
}

\maketitle
% !tex root=./main.tex
\begin{abstract}
Learning from noisy data is a challenging task that significantly degenerates the model performance.
In this paper, we present \methodname, a novel twin contrastive learning model to learn robust representations and handle noisy labels for classification.
Specifically, we construct a Gaussian mixture model (GMM) over the representations by injecting the supervised model predictions into GMM to link label-free latent variables in GMM with label-noisy annotations.
Then, \methodname detects the examples with wrong labels as the out-of-distribution examples by another two-component GMM, taking into account the data distribution.
We further propose a cross-supervision with an entropy regularization loss that bootstraps the true targets from model predictions
to handle the noisy labels.
As a result, \methodname can learn discriminative representations aligned with estimated labels through mixup and contrastive learning.
Extensive experimental results on several standard benchmarks and real-world datasets demonstrate the superior performance of \methodname. In particular, \methodname achieves 7.5\% improvements on CIFAR-10 with 90\% noisy label---an extremely noisy scenario. 
The source code is available at \url{https://github.com/Hzzone/TCL}.
\end{abstract}
% !tex root=./main.tex
\section{Introduction}
Deep neural networks have shown exciting performance for classification tasks~\cite{he2016deep}. Their success largely results from the large-scale curated datasets with clean human annotations, such as CIFAR-10~\cite{krizhevsky2009learning} and ImageNet~\cite{deng2009imagenet}, in which the annotation process, however, is tedious and cumbersome. In contrast, one can easily obtain datasets with some noisy annotations---from online shopping websites~\cite{xiao2015learning}, crowdsourcing~\cite{yan2014learning,yu2018learning}, or Wikipedia~\cite{Rothe-IJCV-2018}---for training a classification neural network. Unfortunately, the mislabelled data are prone to significantly degrade the performance of deep neural networks. Therefore, there is considerable interest in training noise-robust classification networks in recent years~\cite{zheltonozhskii2022contrast,reed2014training,li2020dividemix,li2021learning,ortego2021multi,liu2020early}.

To mitigate the influence of noisy labels, most of the methods in  literature propose the robust loss functions~\cite{zhang2018generalized,wang2019symmetric}, reduce the weights of noisy labels~\cite{xia2019anchor,tanno2019learning}, or correct the noisy labels~\cite{reed2014training,li2020dividemix,ortego2021multi}.
In particular, label correction methods have shown great potential for better performance on the dataset with a high noise ratio. Typically, they correct the labels by using the combination of noisy labels and model predictions~\cite{reed2014training}, which usually require an essential iterative sample selection process~\cite{arazo2019unsupervised,li2020dividemix,ortego2021multi,li2021learning}. 
For example, Arazo~\etal~\cite{arazo2019unsupervised} uses the small-loss trick to carry out sample selection and correct labels via the weighted combination.
In recent years, contrastive learning has shown promising results in handling noisy labels~\cite{ortego2021multi,li2021learning,li2021learning}. They usually leverage contrastive learning to learn discriminative representations, and then clean the labels~\cite{ortego2021multi,li2021learning} or construct the positive pairs by introducing the information of nearest neighbors in the embedding space. However, using the nearest neighbors only considers the label noise within a small neighborhood, which is sub-optimal and cannot handle extreme label noise scenarios, as the neighboring examples may also be mislabeled at the same time. 

To address this issue, this paper presents \methodname, a novel twin contrastive learning model that explores the \emph{label-free} unsupervised representations and \emph{label-noisy} annotations for learning from noisy labels.
Specifically, we leverage contrastive learning to learn discriminative image representations in an unsupervised manner and construct a Gaussian mixture model~(GMM) over its representations. Unlike unsupervised GMM, \methodname links the \emph{label-free} GMM and  \emph{label-noisy} annotations by replacing the latent variable of GMM with the model predictions for updating the parameters of GMM.
Then, benefitting from the learned data distribution, we propose to formulate label noise detection as an out-of-distribution~(OOD) problem, utilizing another two-component GMM to model the samples with clean and wrong labels. The merit of the proposed OOD label noise detection is to take the full data distribution into account, which is robust to the neighborhood with strong label noise.
Furthermore, we propose a bootstrap cross-supervision with an entropy regulation loss to reduce the impact of wrong labels, in which the true labels of the samples with wrong labels are estimated from another data augmentation.
Last, to further learn robust representations, we leverage contrastive learning and Mixup techniques to inject the structural knowledge of classes into the embedding space, which helps align the representations with estimated labels.

The contributions are summarized as follows:
\begin{itemize}
\setlength{\itemsep}{0pt}%
\item We present \methodname, a novel twin contrastive learning model that explores the \emph{label-free} GMM for unsupervised representations and \emph{label-noisy} annotations for learning from noisy labels. 

\item We propose a novel OOD label noise detection method by modeling the data distribution, which excels at handling extremely noisy scenarios.

\item We propose an effective cross-supervision, which can bootstrap the true targets with an entropy loss to regularize the model.

\item Experimental results on several benchmark datasets and real-world datasets demonstrate that our method outperforms the existing state-of-the-art methods by a significant margin. In particular, we achieve 7.5\% improvements in extremely noisy scenarios.
\end{itemize}
% !tex root=./main.tex
\section{Related Work}
\paragraph{Contrastive learning.}
Contrastive learning methods~\cite{wu2018unsupervised,he2020momentum,chen2020simple} have shown promising results for both representation learning and downstream tasks. The popular loss function is InfoNCE loss~\cite{oord2018representation}, which can pull together the data augmentations from the same example and push away the other negative examples. MoCo~\cite{he2020momentum} uses a memory queue to store the consistent representations. SimCLR~\cite{chen2020simple} optimizes InfoNCE within mini-batch and has found some effective training tricks, \eg, data augmentation.
However, as unsupervised learning, they mainly focus on inducing transferable representations for the downstream tasks instead of training with noisy annotations. Although supervised contrastive learning~\cite{khosla2020supervised} can improve the representations by human labels, it harms the performance when label noise exists~\cite{li2022selective}.

\paragraph{Learning with noisy labels.}
Most of the methods in literature mitigate the label noise by robust loss functions~\cite{xu2019l_dmi,wang2019symmetric,zhang2018generalized,wang2019imae,ghosh2017robust,liu2020early}, noise transition matrix~\cite{goldberger2016training,patrini2017making,xia2019anchor,tanno2019learning}, sample selection~\cite{han2018co,yu2019does}, and label correction~\cite{reed2014training,tanaka2018joint,li2020dividemix,li2020mopro,li2021learning,kim2021fine,ortego2021multi,liu2020early}. In particular, label correction methods have shown promising results than other methods.
Arazo~\etal~\cite{arazo2019unsupervised} applied a mixture model to the losses of each sample to distinguish the noisy and clean labels, inspired by the fact that the noisy samples have a higher loss during the early epochs of training.
Similarly, DivideMix~\cite{li2020dividemix} employs two networks to perform the sample selection for each other and applies the semi-supervised learning technique where the targets are computed from the average predictions of different data augmentations.
Due to the success of contrastive learning, many attempts have been made to improve the robustness of classification tasks by combining the advantages of contrastive learning. Zheltonozhskii~\etal~\cite{zheltonozhskii2022contrast} used contrastive learning to pre-train the classification model. MOIT~\cite{ortego2021multi} quantifies this agreement between feature representation and original label to identify mislabeled samples by utilizing a $k$-nearest neighbor ($k$-NN) search. RRL~\cite{li2021learning} performs label cleaning by two thresholds on the soft label, which is calculated from the predictions of previous epochs and its nearest neighbors. Sel-CL~\cite{li2022selective} leverages the nearest neighbors to select confident pairs for supervised contrastive learning~\cite{khosla2020supervised}.

Unlike existing methods~\cite{ortego2021multi,li2021learning,li2022selective} that detect the wrong labels within the neighborhood, \methodname formulates the wrong labels as the out-of-distribution examples by modeling the data distribution of representations learned by contrastive learning. In addition, we propose a cross-supervision with entropy regularization to better estimate the true labels and handle the noisy labels.

\begin{figure*}[t]
    \centering
    \includegraphics[width=1.0\linewidth]{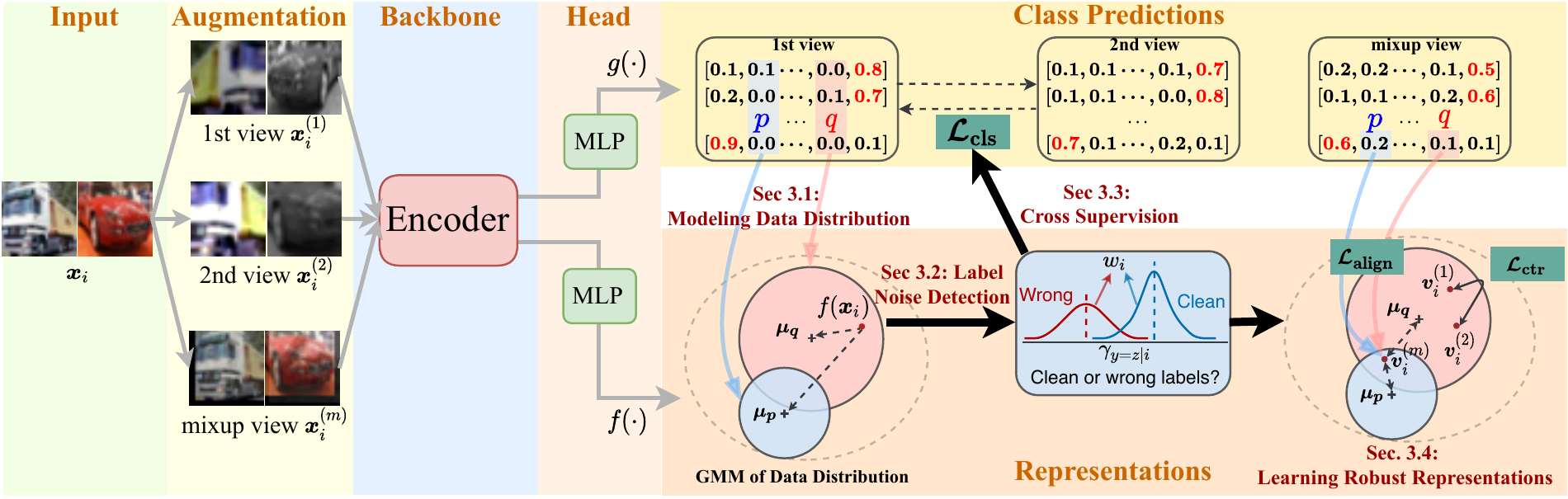}
    \caption{
    Illustration of the proposed \methodname.
    The networks $g$ and $f$ with shared encoder and independent two-layer MLP output the class predictions and representations. Then, \methodname models the data distribution via a GMM, and detects the examples with wrong labels as out-of-distribution examples. To optimize~\methodname, these results lead to cross-supervision and robust representation learning.
    \label{fig:framework}
    }
\end{figure*}
% !tex root=./main.tex
\section{The Proposed \methodname}

Each image in dataset $\mathcal{D}=\{\mat{x}_i\}_{i=1}^N$ associates with an annotation $y\in\{1, 2,\ldots, K\}$. In practice, some examples may be mislabeled. We aim to train a classification network, $p_{\theta}(y|\vct{x}) = g(\mat{x};\theta) \in \mathbb{R}^{K}$, that is resistant to the noisy labels in training data, and generalizes well on the clean testing data.
Fig.~\ref{fig:framework} illustrates the framework of our proposed \methodname. 

\paragraph{Overview.}
In the context of our framework, $f(\cdot)$ and $g(\cdot)$ share the same backbone and have additional individual heads to output representations and class predictions from two random and one mixup data augmentations. Afterward, there are four components in \methodname, including (i) modeling the data distribution via a GMM in Sec.~\ref{sec:model_data} from the model predictions and representations; (ii) detecting the examples with wrong labels as out-of-distribution samples in Sec.~\ref{sec:noise_detction}; (iii) cross-supervision by bootstrapping the true targets in Sec.~\ref{sec:cross_supervision}; and (iv) learning robust representations through contrastive learning and mixup in Sec.~\ref{sec:robust_representation}.

\subsection{Modeling Data Distribution}
\label{sec:model_data}
Given the image dataset consisting of $N$ images, we opt to model the distribution of $\vct{x}$ over its representation $\vct{v} = f(\vct{x})$ via a spherical Gaussian mixture model (GMM).
After introducing discrete latent variables $z\in\{1, 2,\ldots, K\}$ that determine the assignment of observations to mixture components, the unsupervised GMM can be defined as
\begin{align}
    p(\vct{v} ) &= \sum\nolimits_{k=1}^K p(\vct{v}, z=k) \notag\\ 
    &=\sum\nolimits_{k=1}^K p(z=k) \mathcal{N}(\vct{v} | \vct{\mu}_k, \sigma_k).\label{eq:gmm}
\end{align}
where $\vct{\mu}_k$ is the mean and $\sigma_k$ a scalar deviation.
If we assume that the latent variables $z$ are uniform distributed, that is, $p(z=k) = 1/K$, we can define the posterior probability that assigns $\vct{x}_i$ to $k$-th cluster:
\begin{align}
\gamma_{ik}=p(z_i=k|\vct{x}_i)\propto \mathcal{N}(\vct{x}_i|\vct{\mu}_k,\sigma_k).\label{eq:posterior_probability}
\end{align}
In an ideal scenario where all the samples have clean labels $y\in \{1, 2,\ldots, K\}$, the discrete latent variables $z$ would be identical to the annotation $y$, and the parameters $\vct{\mu}_k$, $\sigma_k$ and latent variable $z$ can be solved through a standard Expectation-Maximization (EM) algorithm~\cite{dempster1977maximum}.

However, in practice, the labels are often noisy and the latent variable $z$, estimated in an unsupervised manner, has nothing to do with the label $y$. Therefore, we are interested in connecting latent variable $z$, estimated \emph{in an unsupervised fashion  (\ie~label-free)}, and the available annotations $y$, \emph{label-noisy}, for the task of learning from noisy labels. 

To link them together, we propose to inject the model predictions $p_{\theta}(y_i = k|\vct{x}_i)$, learned from noisy labels, into the latent variables $z$. Specifically, we propose to replace the unsupervised assignment $p(z_i=k|\vct{x}_i)$ with noisy-supervised assignment $p_\theta(y_i=k|\vct{x}_i)$. As a result, we can connect the latent variable $z$ with the label $y$, and thus use the noisy supervision to guide the update of the parameters of GMM. In particular, the update of the GMM parameters becomes
\begin{align}
     \vct{\mu}_k &= \operatorname{norm}\left(\frac{\sum_i p_{\theta}(y_i = k|\vct{x}_i) \vct{v}_i}{\sum_i p_{\theta}(y_i = k|\vct{x}_i)} \right),
     \\ 
     \sigma_k &= \frac{\sum_i p_{\theta}(y_i = k|\vct{x}_i) (\vct{v}_i - \vct{\mu}_k)(\vct{v}_i - \vct{\mu}_k)\T}{\sum_i p_{\theta}(y_i = k|\vct{x}_i)},
\end{align}
where $\operatorname{norm}(\cdot)$ is $\ell_2$-normalization such that $\|\vct{\mu}_k\|_2=1$.

\subsection{Out-Of-Distribution Label Noise Detection}
\label{sec:noise_detction}

Previous works~\cite{ortego2021multi,li2021learning,li2022selective} typically detect the wrong labels within the neighborhood, that is, using the information from nearest neighbors. It is limited as the neighboring examples are usually mislabeled at the same time. To address this issue, we propose to formulate label noise detection as to detect the out-of-distribution examples.

After building the connection between the latent variables $z$ and labels $y$, we are able to detect the sample with wrong labels through the posterior probability in Eq.~\eqref{eq:posterior_probability}. We implement it as a  normalized version to take into account the intra-cluster distance, which allows for detecting the samples with likely wrong labels:
\begin{align}
\gamma_{ik}=\frac{\exp\left(- (\vct{v}_i-\vct{\mu}_k)\T(\vct{v}_i-\mu_k) / 2\sigma_k\right)}{\sum_k \exp\left(- (\vct{v}_i-\vct{\mu}_k)\T(\vct{v}_i-\vct{\mu}_k) / 2\sigma_k\right)}.\label{eq:posterior_normalized}
\end{align}
Since $\ell_2$-normalization has been applied to both embeddings $\vct{v}$ and the cluster centers $\vct{\mu}_k$, yielding $(\vct{v}-\vct{\mu}_k)\T(\vct{v}-\vct{\mu}_k)=2-2\vct{v}\T \vct{\mu}_k$. Therefore, we can re-write Eq.~\eqref{eq:posterior_normalized} as:
\begin{align}
    \gamma_{ik}&=p(z_i=k|\vct{x}_i)\notag\\
    &=\exp(\vct{v}_i\T\vct{\mu}_k / \sigma_k) \Big/ \sum\nolimits_k \exp(\vct{v}_i\T \vct{\mu}_k/\sigma_k).\label{eq:posterior_normalized_sim}
\end{align}

Once built the GMM over the distribution of representations, we propose to formulate the conventional \emph{noisy label} detection problem  as \emph{out-of-distribution sample} detection problem. Our idea is that the samples with clean labels should have the same cluster indices after linking the cluster index and class label. Specifically, given one particular class $y=k$, the samples within this class can be divided into two types: in-distribution samples with clean labels, and out-of-distribution samples with wrong labels. Therefore, we define the following conditional probability to measure the probability of one sample with clean label:
\begin{align}
    \gamma_{y=z|i}&=p(y_i=z_i|\vct{x}_{i})\notag\\
    &=\exp(\vct{v}_i\T \vct{\mu}_{z_i} / \sigma_{z_i}) \Big/ \sum\nolimits_k \exp(\vct{v}_i\T \vct{\mu}_k/\sigma_k).\label{eq:ood}
\end{align}
Although Eqs.~\eqref{eq:posterior_normalized_sim} and~\eqref{eq:ood} share similar calculations, they have different meanings. Eq.~\eqref{eq:posterior_normalized_sim} calculates the probability of one example \emph{belonging to $k$-th cluster} while Eq.~\eqref{eq:ood} the probability of one example \emph{having clean label}---that is, $y_i=z_i$. Therefore, the probability of one example having the wrong label can be written as $\gamma_{y\neq z|i}= p(y_i\neq z_i|\vct{x}_{i}) = 1 - p(y_i=z_i|\vct{x}_{i})$. 

Furthermore, instead of setting a human-tuned threshold for $\gamma_{y=z|i}$, we opt to employ another two-component GMM following~\cite{arazo2019unsupervised,li2020dividemix} to automatically estimate the clean probability $\gamma_{y=z|i}$ for each example. Similar to the definition of GMM in Eq.~\eqref{eq:gmm}, this two-components GMM is defined as follows:
\begin{align}
    p(\gamma_{y=z|i})=\sum_{c=0}^{1}p(\gamma_{y=z|i}, c) = \sum_{c=0}^{1} p(c)p(\gamma_{y=z|i}| c),\label{eq:gmm_two}
\end{align}
where $c$ is the new introduced latent variable: $c=1$ indicates the cluster of clean labels with higher mean value and vice versus $c=0$. After modeling the GMM over the probability of one example having clean labels, $\gamma_{y=z|i}$, we are able to infer the posterior probability of one example having clean labels through the two-component GMM.

\subsection{Cross-supervision with Entropy Regularization}
\label{sec:cross_supervision}
After the label noise detection, the next important step is to estimate the true targets by correcting the wrong label to reduce its impact, called label correction.
Previous works usually perform label correction using the temporal ensembling~\cite{liu2020early} or from the model predictions~\cite{arazo2019unsupervised,li2020dividemix} before mixup augmentation without back-propagation.

\methodname leverages a similar idea to bootstrap the targets through the convex combination of its noisy labels and the predictions from the model itself:
\begin{align}
\begin{cases}
    \vct{t}^{(1)}_i = w_i \vct{y}_i + (1 - w_i) g(\mat{x}^{(1)}_i)\\
    \vct{t}^{(2)}_i = w_i \vct{y}_i + (1 - w_i) g(\mat{x}^{(2)}_i) 
    \end{cases},
    \label{eq:convex_comb}
\end{align}
where $g(\mat{x}^{(1)}_i)$ and $g(\mat{x}^{(2)}_i)$ are the predictions of two augmentations, $\vct{y}_i$ the noisy one-hot label, and $w_i\in[0, 1]$ represents the posterior probability as $p(c=1|\gamma_{y=z|i})$ from the two-component GMM defined in Eq.~\eqref{eq:gmm_two}. When computing Eq.~\eqref{eq:convex_comb}, we stop the gradient from $g$ to avoid the model predictions collapsed into a constant, inspired by~\cite{chen2021exploring,grill2020bootstrap}.

Guided by the corrected labels $\vct{t}_i$, we swap two augmentations to compute the classification loss twice, leading to the bootstrap cross supervision, formulated as:
\begin{align}
    \mathcal{L}_{\mathrm{cross}} = \ell\left(g(\mat{x}_i^{(1)}), \vct{t}^{(2)}_i\right) + \ell\left(g(\mat{x}_i^{(2)}), \vct{t}^{(1)}_i\right),\quad 
\end{align}
where $\ell$ is the cross-entropy loss. This loss makes the predictions of the model from two data augmentations close to corrected labels from each other. In a sense, if $w_i=0$, the model is encouraged for consistent class predictions between different data augmentations, otherwise $w_i=1$ it is supervised by the clean labels.

In addition, we leverage an additional entropy regularization loss on the predictions within a mini-batch $\mathcal{B}$:
\begin{align}
    \mathcal{L}_{\mathrm{reg}} &= - \mathbb{H}\left(\frac{1}{|\mathcal{B}|}\sum_{\mat{x}\in \mathcal{B}}  g(\mat{x})\right) +\frac{1}{|\mathcal{B}|}\sum_{\mat{x}\in \mathcal{B}} \mathbb{H}\left(g(\mat{x})\right),
\end{align}
where $\mathbb{H}(\cdot)$ is the entropy of predictions~\cite{shannon2001mathematical}. The first term can avoid the predictions collapsing into a single class by maximizing the entropy of average predictions.
The second term is the minimum entropy regularization to encourage the model to have high confidence for predictions, which was previously studied in semi-supervised learning literature~\cite{grandvalet2004semi}.

Although both using the model predictions, we would emphasize that the cross-supervision in \methodname is different to~\cite{liu2020early,arazo2019unsupervised,li2020dividemix} in three aspects: (i) both $\mat{x}^{(1)}_i$ and $\mat{x}^{(2)}_i$ are involved in back-propagation; (ii) the strong augmentation~\cite{chen2020simple} used to estimate the true targets can prevent the overfitting of estimated targets; and (iii) \methodname employs two entropy regularization terms to avoid the model collapse to one class.

The final classification loss is given as follows:
\begin{align}
    \mathcal{L}_{\mathrm{cls}} = \mathcal{L}_{\mathrm{cross}} + \mathcal{L}_{\mathrm{reg}}.
\end{align}

\subsection{Learning Robust Representations}
\label{sec:robust_representation}
To model the data distribution that is robust to noisy labels, we leverage contrastive learning to learn the representations of images. Specifically,
contrastive learning performs instance-wise discrimination~\cite{wu2018unsupervised} using the InfoNCE loss~\cite{oord2018representation} to enforce the model outputting similar embeddings for the images with semantic preserving perturbations.
Formally, the contrastive loss is defined as follows:
\begin{align}
    \mathcal{L}_{\mathrm{ctr}} = -\log \frac{\exp\left( f(\mat{x}^{(1)})\T f(\mat{x}^{(2)}) / \tau\right)}{\sum\nolimits_{{\mat{x}}\in \mathcal{S}} \exp\left(f(\mat{x}^{(1)})\T f({\mat{x}}) / \tau\right)},\label{eq:contrastive_loss}
\end{align}
where $\tau$ is the temperature and $\mathcal{S}$ is the $\mathcal{B}$ except $\mat{x}^{(1)}$. $\mat{x}^{(1)}$ and $\mat{x}^{(2)}$ are two
augmentations of $\mat{x}$.
Intuitively, InfoNCE loss aims to pull together the positive pair ($\mat{x}^{(1)}, \mat{x}^{(2)}$) from two different augmentations of the same instance, and push them away from negative examples of other instances. Consequently, it can encourage discriminative representations in a pure unsupervised, or label-free manner.

Although beneficial in modeling latent representations, contrastive learning cannot introduce compact classes without using the true labels. Since the label $\vct{y}$ is noisy, we leverage Mixup~\cite{zhang2017mixup} to improve within-class compactness, which has been shown its effectiveness against label noise in literature~\cite{arazo2019unsupervised,li2020dividemix}. Specifically,
a mixup training pair $(\mat{x}^{(\mathrm{m})}_i, \bar{\vct{t}}^{(\mathrm{m})}_i)$ is linearly interpolated between $(\mat{x}_i, \bar{\vct{t}}_i)$ and $(\mat{x}_j, \bar{\vct{t}}_j)$ under a control coefficient $\lambda\sim\operatorname{Beta}(\alpha,\alpha)$:
\begin{align}
\begin{cases}
    \mat{x}^{(\mathrm{m})}_i = \lambda \mat{x}_i + (1 - \lambda) \mat{x}_j,\\ \bar{\vct{t}}^{(\mathrm{m})}_i = \lambda \bar{\vct{t}}_i + (1 - \lambda) \bar{\vct{t}}_j,
    \end{cases}
\end{align}
where $\mat{x}_j$ is randomly selected within a mini-batch, and $\bar{\vct{t}}_i=(\vct{t}^{(1)}_i+\vct{t}^{(2)}_i)/2$ is the average of estimated true labels of two data augmentations. Intuitively, we can inject the structural knowledge of classes into the embedding space learned by contrastive learning. This loss can be written as:
\begin{align}
    \mathcal{L}_{\mathrm{align}} = \ell\left(g(\mat{x}^{(\mathrm{m})}_i), \bar{\vct{t}}^{(\mathrm{m})}_i\right) + \ell(p(\vct{z}|\vct{x}^{(\mathrm{m})}_i), \bar{\vct{t}}^{(\mathrm{m})}_i),
\end{align}
where the second term can align the representations with estimated labels. In a sense, $\mathcal{L}_{\mathrm{align}}$ regularizes classification network $g$ and encourages $f$ to learn compact and well-separated representations.
Furthermore, we would point out two differences between \methodname and~\cite{li2021learning}, although both using mixup to boost the representations. First,~\cite{li2021learning} does not explicitly model the data distribution $p(\vct{z}|\vct{x}^{(\mathrm{m})}_i)$ like \methodname. Second, \methodname has leveraged the full training dataset via the corrected label $\bar{\vct{t}}^{(\mathrm{m})}_i$ instead of a subset of clean examples in~\cite{li2021learning}, which leads to stronger robustness of \methodname over~\cite{li2021learning} on extreme high label noise ratios.

\subsection{Training and inference}
\label{sec:training}

\begin{algorithm}[t]
        \caption{Training Algorithm}\label{alg}
        \textbf{Input:}\hspace{0mm} Dataset $\mathcal{D}=\{(\mat{x}_i, y_i)\}_{i=1}^N$; functions $\{f, g\}$ \\%and the max iteration $T$.\\
        \textbf{Output:}\hspace{0mm} Classification network $g$.\\
        \Repeat{reaching max epochs}{
        E-step: update $\{(\vct{\mu}_k, \sigma_k)\}_{k=1}^K$ for \methodname, and $\{w_i\}_{i=1}^N$ for each sample in $\mathcal{D}$ \\
        M-step:
        \Repeat{an epoch finished}{
            Randomly sample a mini-batch $\mathcal{B}$ from $\mathcal{D}$\\
            \For{each $\mat{x}_i$ in $\mathcal{B}$}{
                Randomly sample two augmentations and a mixup one: \{$\mat{x}^{(1)}_i$, $\mat{x}^{(2)}_i$, $\mat{x}_i^{(\mathrm{m})}$\} \\
                $\mathcal{L}\leftarrow$Eq.~\eqref{eq:final_loss}\\
            }
            Update $f$ and $g$ with SGD optimizer. \\
        }
        }
    \end{algorithm}

The overall training objective is to minimize the sum of all losses:
\begin{align}
    \mathcal{L} = \mathcal{L}_{\mathrm{cls}} + \mathcal{L}_{\mathrm{ctr}} + \mathcal{L}_{\mathrm{align}}.
    \label{eq:final_loss}
\end{align}
We find that a simple summation of all losses works well for all datasets and noise levels, which indicates the strong generalization of the proposed method. During inference, the data augmentations are disabled and the class predictions are obtained by $\operatorname{argmax}_{k} p_{\theta}(k |\mathbf{x})$.

The training algorithm of the proposed method is shown in Alg.~\ref{alg}. In a sense, the architecture of our method leads to an EM-like algorithm: (1) the \bd{E-step} updates $\{(\vct{\mu}_k, \sigma_k)\}_{k=1}^K$ for \methodname, and $\{w_i\}_{i=1}^N$ for each sample in $\mathcal{D}$ to form the true targets with the predictions from another data augmentations, and (2) the \bd{M-step} optimizes the model parameters by Eq.~\eqref{eq:final_loss} to better fit those estimated targets.
Therefore, the convergence of \methodname can be theoretically guaranteed, following the standard EM algorithm. 
% !tex root=./main.tex
\section{Experiments}
\label{sec:exp}
In this section, we conduct experiments on multiple benchmark datasets with simulated and real-world label noises. We strictly follow the experimental settings in previous literature~\cite{ortego2021multi,li2020dividemix,li2021learning,liu2020early} for fair comparisons.

\subsection{Experiments on simulated datasets}

\paragraph{Datasets.} 
Following~\cite{ortego2021multi,li2020dividemix,li2021learning,liu2020early}, we validate our method on \bd{CIFAR-10/100}~\cite{krizhevsky2009learning}, which contains 50K and 10K images with size $32\times 32$ for training and testing, respectively. We leave 5K images from the training set as the validation set for hyperparameter tuning, then train the model on the full training set for fair comparisons.
Two types of label noise are simulated: \emph{symmetric} and \emph{asymmetric} label noise. Symmetric noise randomly assigns the labels of the training set to random labels with predefined percentages, \textit{a.k.a}, noise ratio, which includes 20\%, 50\%, 80\%, and 90\% on two datasets in this paper. Asymmetric noise takes into account the class semantic information, and the labels are only changed to similar classes (\eg, truck $\to$ automobile). Here, only experiments on the CIFAR-10 dataset with 40\% noise ratio for asymmetric noise are conducted; otherwise, the classes with above $50\%$ label noise cannot be distinguished.

\begin{table*}[t]
    \centering
    \caption{
        Comparisons with state-of-the-art methods on simulated datasets.
        The results for previous methods are copied from~\cite{li2021learning,li2022selective} to avoid the bias of self-implementation, and we strictly follow their experimental settings.
        Each runs has been repeated 3 times with different randomly-generated noise and we report the mean and std values of \emph{last } 5 epochs.
    }
    \label{tab:results_on_cifar}
    \begin{tabular}{rrrrrrrrcrrrr}
    
    & \multicolumn{6}{c}{\bd{CIFAR-10}} & & \multicolumn{5}{c}{\bd{CIFAR-100}} \\
    \multirow{2}{*}{Noise type/rate} & \multicolumn{4}{c}{\textit{Sym.}} & \textit{Asym.} & \textit{Avg.} & & \multicolumn{4}{c}{\textit{Sym.}} & \textit{Avg.} \\ 
    \cmidrule{2-7}\cmidrule{9-13}
    & 20\% & 50\% & 80\% & 90\% & 40\% & & &20\%  & 50\%  & 80\%  & 90\% & \\
    \shline
    Cross-Entropy        & 82.7  & 57.9  & 26.1  & 16.8  & 76.0 & 51.9 & & 61.8  & 37.3  & 8.8 & 3.5 &  27.8 \\
    Mixup~(17')~\cite{zhang2017mixup} & 92.3  & 77.6  & 46.7  & 43.9  & 77.7 & 67.6 & & 66.0  & 46.6  & 17.6  & 8.1 & 34.6 \\
    P-correction~(19')~\cite{yi2019probabilistic} & 92.0  & 88.7  & 76.5  & 58.2  & 91.6 & 81.4 && 68.1  & 56.4  & 20.7  & 8.8 & 38.5 \\
    M-correction~(19')~\cite{arazo2019unsupervised} & 93.8  & 91.9  & 86.6  & 68.7  & 87.4 & 85.7 && 73.4  & 65.4  & 47.6  & 20.5 & 51.7 \\
    ELR~(20')~\cite{liu2020early} & 93.8  & 92.6  & 88.0  & 63.3  & 85.3  & 84.6 && 74.5  & 70.2  & 45.2  & 20.5  & 52.6 \\
    DivideMix~(20')~\cite{li2020dividemix} & \underline{95.0} & 93.7  & \underline{92.4} & 74.2  & 91.4  & 89.3 && 74.8 & 72.1  & 57.6 & 29.2 &  58.4 \\
    MOIT~(21')~\cite{ortego2021multi} & 93.1  & 90.0 & 79.0 & 69.6 & 92.0 & 84.7& & 73.0 & 64.6 & 46.5 & 36.0 & 55.0 \\
    RRL~(21')~\cite{li2021learning} & \bd{95.8}  & \bd{94.3} & \underline{92.4} & 75.0  & 91.9 & 89.8 && \bd{79.1} & \bd{74.8}  & 57.7 & 29.3  & 60.2 \\
    Sel-CL+~(22')~\cite{li2022selective} & \underline{95.5}  & \underline{93.9} & 89.2 & \underline{81.9}  & \bd{93.4} & \underline{90.7} && 76.5 & 72.4  & \underline{59.6} & \underline{48.8}  & \underline{64.3} \\
    \hline
    \methodname (\bd{ours}) & 95.0\ & \underline{93.9} & \bd{92.5} & \bd{89.4} & \underline{92.6} & \bd{92.7} && \underline{78.0} & \underline{73.3} & \bd{65.0} & \bd{54.5} & \bd{67.7} \\
     & \std{0.1}\ & \std{0.1} & \std{0.2} & \std{0.2} & \std{0.1} &  && \std{0.2} & \std{0.2} & \std{0.3} & \std{0.5} & \\
    \end{tabular}
\end{table*}

\begin{figure*}[t]
    \centering
    \includegraphics[width=1.0\linewidth]{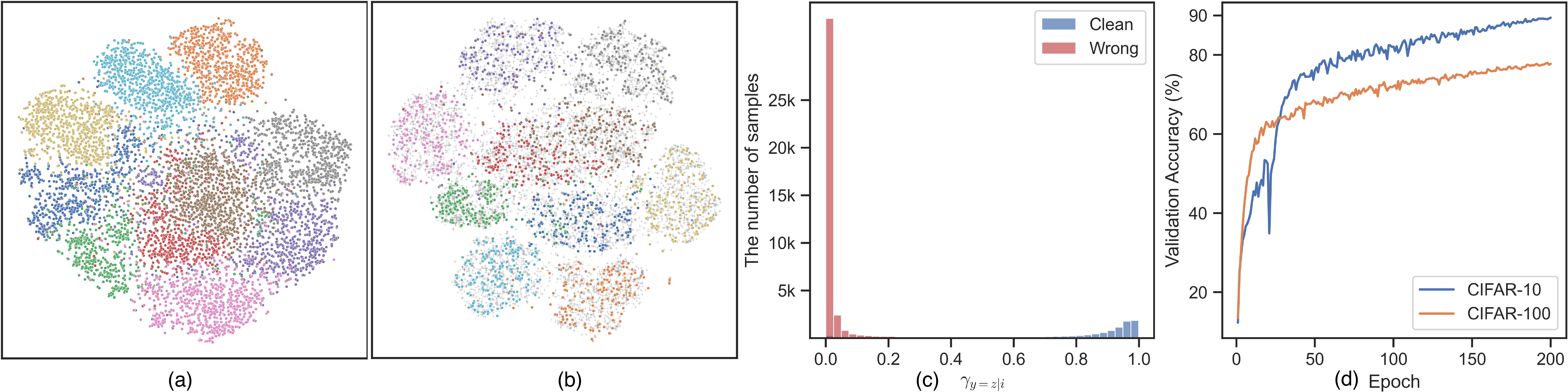}
    \caption{
        Qualitative results.
        For the model trained on CIFAR-10 with 90\% \textit{sym.} noise at 200th epoch, we show t-SNE visualizations for the learned representations of (a) testing set where different color denotes different class predicted by $g(\cdot)$ and (b) 10K samples from training set colored by the true labels; the gray `+' denotes the samples with noisy labels.  (c) The histogram of $p(y=z|\vct{x})$ for full training set colored by the clean and noisy labels. (d) The validation accuracy across training of CIFAR-10 and CIFAR-100 on 90\% \textit{sym.} noise.
    }
    \label{fig:qualitive_results}
\end{figure*}

\paragraph{Training details.} 
Same as previous works~\cite{ortego2021multi,li2020dividemix,li2021learning,liu2020early}, we use a PreAct ResNet-18~\cite{he2016identity} as the encoder. We adopt SGD optimizer to train our model with a momentum of 0.9, a weight decay of 0.001, and a batch size of 256 for 200 epochs. The learning rate are linearly warmed up to 0.03 for 20 epochs and decayed with the cosine schedule.
The data augmentations of~\cite{chen2020simple} are applied to two views~(ResizedCrop, ColorJitter, and \textit{etc}). Only crop and horizontal flip are employed for mixup. Both projection and classification heads are a two-layer MLP with the dimension 128 and the number of classes. The temperature $\tau$ of contrastive loss and the $\alpha$ of mixup are 0.25 and 1. The settings are shared for all experiments, which are significantly different from~\cite{li2020dividemix,li2021learning,liu2020early} that adopt specific configurations for different datasets and even for different noise ratios/types.

\paragraph{Quantitative results.}
Table~\ref{tab:results_on_cifar} presents the the comparisons with existing state-of-the-art methods. Our method yields competitive performance on low noise ratios, but promising improvements over recent methods on extreme noise ratios and the most challenging CIFAR-100 dataset with 100 classes. In particular, with 90\% label noise, there are 7.5\% and 5.7\% improvements on for CIFAR-10 and CIFAR-100, respectively. We stress that the hyperparameters are consistent for different noise ratios/types. In practical scenarios, the noise ratio for a particular dataset is unknown, so it is hard to tune the hyper-parameters for better performance. Therefore, these results indicate the strong generalization ability of our method regardless of noise ratios/types.

For fair comparisons, following~\cite{ortego2021multi,li2022selective}, we performed extra experiments on fine-tuning the classification network for 70 epochs with the detected clean samples and mixup augmentation, termed TCL+. Table~\ref{tab:results_on_cifar_plus} shows that under low label noise~(below 50\%), TCL+ achieves significant improvements over TCL and outperforms the recent state-of-the-art methods. The benefits from the detected clean subset and longer training, which can fully utilize the useful supervision signals from labeled examples.

In Appendix~\ref{sec:knn}, we also perform the $k$-NN classification over the learned representations, which indicates that our method has maintained meaningful representations better than the pure unsupervised learning model.
In Appendix~\ref{sec:asym_noise}, we provide more experimental results and analysis on asymmetric label noise and imbalance data.

\begin{table}[t]
   \centering
   \scalebox{0.95}{
   \begin{tabular}{rrrrrrr}
   & \multicolumn{3}{c}{\bd{CIFAR-10}} & & \multicolumn{2}{c}{\bd{CIFAR-100}} \\
   & \multicolumn{2}{c}{\textit{Sym.}} & \textit{Asym.} & & \multicolumn{2}{c}{\textit{Sym.}} \\ 
   \cmidrule{2-4} \cmidrule{6-7}
   & 20\% & 50\% & 40\% & &20\%  & 50\% \\
   \shline
   DivideMix~\cite{li2020dividemix} & 95.0 & 93.7  & 91.4 & & 74.8 & 72.1  \\
   MOIT~\cite{ortego2021multi} & 93.1  & 90.0 & 92.0 & & 73.0 & 64.6 \\
   MOIT+~\cite{ortego2021multi} & 94.1  & 91.8 & 93.3 & & 75.9 & 70.6 \\
   RRL~\cite{li2021learning} & \underline{95.8} & \underline{94.3} & 91.9 & & \underline{79.1} & \bd{74.8}  \\
   Sel-CL+~\cite{li2022selective} & 95.5  & 93.9 & \underline{93.4} & & 76.5 & 72.4 \\
   \hline
   \methodname (\bd{ours}) & 95.0\ & 93.9 &  92.6 & & 78.0 & 73.3 \\
   TCL+ (\bd{ours}) & \bd{96.0} & \bd{94.5} & \bd{93.7} & & \bd{79.3} & \underline{74.6} \\
   \shline
   \end{tabular}
   }
   \caption{
      Comparisons with SOTAs under \emph{low} label noise.
   }
   \label{tab:results_on_cifar_plus}
\end{table}

\paragraph{Qualitative results.} Figs.~\ref{fig:qualitive_results}(a) and (b) visualize the learned representations with extremely high noise ratio, demonstrating that our method can produce distinct structures of learned representations with meaningful semantic information. Especially, Fig.~\ref{fig:qualitive_results}(b) presents the samples with noisy labels in the embedding space, in which the label noise can be accurately detected by our proposed method. In addition, by visualizing the histogram of $p(y=z|\vct{x})$ for the training set in Fig.~\ref{fig:qualitive_results}(c), we confirm that the proposed method can effectively distinguish the samples noisy and clean labels.
We visualize the validation accuracy across training in Fig.~\ref{fig:qualitive_results}(d). As expected, \methodname performs stable even with the extreme 90\% label noise.

\subsection{Ablation study}

We conduct ablation studies to validate our motivation and design with the following baselines, and the results are shown in Table~\ref{tab:ablation_study}.

\begin{enumerate}[label=\color{red!70!black}(\roman*),wide,labelindent=0pt,itemsep=0ex,parsep=0pt,topsep=0pt]
\item\label{bl_1} \textbf{Baseline.}\quad
We start the baseline method by removing the proposed noisy label detection and bootstrap cross-supervision, where the model is directly guided by noisy labels.  As expected, the performance significantly degrades for the extremely high noise ratio~(\ie, 90\%).

\item\label{bl_2} \textbf{Label Noise Detection.}\quad
We assess the effectiveness of different detection methods including the cross-entropy loss~\cite{arazo2019unsupervised,li2020dividemix}, $k$-NN search~\cite{ortego2021multi}, and our out-of-distribution~(OOD) detection. For fair comparisons, the predictions from the images before mixup are employed as the true labels in Eq.~\eqref{eq:convex_comb}.
Obviously, the label noise detection has alleviated the degeneration to some degree~(Exp.~\ref{bl_1}), where our method consistently outperforms other baselines.
Fig.~\ref{fig:label_noise_detection_AUC} visualizes their AUCs across training.
The proposed OOD detection is better at distinguishing clean and wrong labels. Thanks to the representations learned by contrastive learning, $k$-NN search performs better than cross-entropy loss. However, it is limited due to the use of the original labels to detect noisy ones, while our method constructs a GMM using the model predictions.

\begin{figure}[t]
    \centering
    \includegraphics[width=0.75\linewidth]{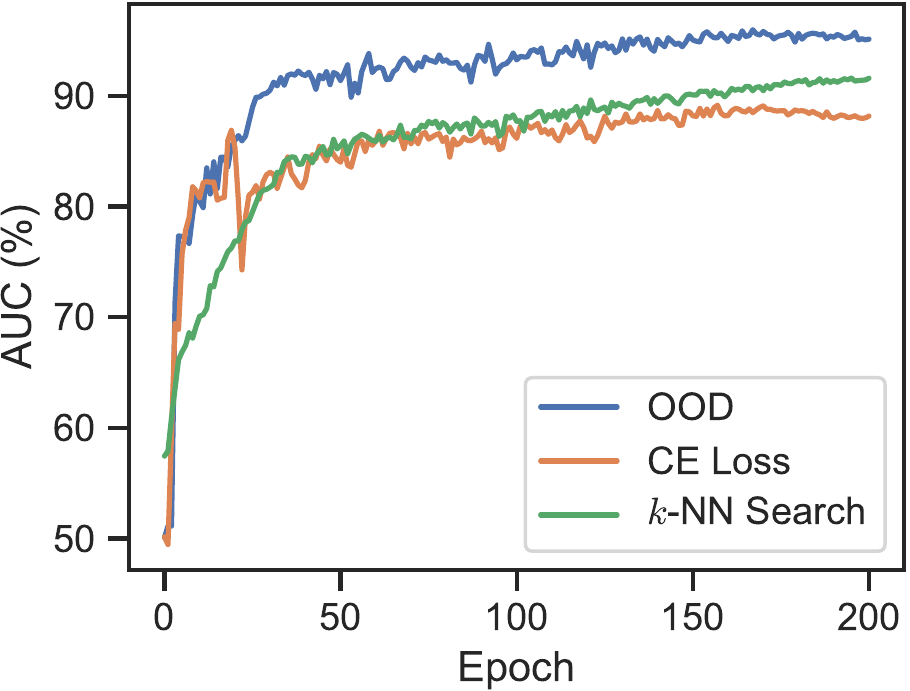}
    \caption{
        Training curve of AUC for noisy label detection trained on CIFAR-10 with 90\% \textit{sym.} noise. 
    }
    \label{fig:label_noise_detection_AUC}%
\end{figure}

\item\label{bl_3} \textbf{Target Estimation.}\quad
Another key component is the cross-supervision that bootstraps the true targets from the predictions of another data augmentation. We replace it with the temporal ensembling~\cite{liu2020early}, where the hyperparameters are set as suggested by~\cite{liu2020early}. Furthermore, Exp.~\ref{bl_3} estimates true targets from the images before mixup~\cite{li2020dividemix,liu2020early,arazo2019unsupervised}. The results suggest that our bootstrap cross-supervision has shown strong robustness on 90\% label noise.

\begin{table}[t]
    \centering
    \scalebox{.73}{
    \begin{tabular}{llcccccccc}
    \shline
    \multicolumn{2}{l}{Dataset}   & \multicolumn{4}{c}{\bd{CIFAR-10}} && \multicolumn{3}{c}{\bd{CIFAR-100}} \\
    \midrule
    \multicolumn{2}{l}{\multirow{3}{*}{Noise type/rate}} & \multicolumn{2}{c}{\textit{Sym.}} & \textit{Asym.} & \textit{Avg.} && \multicolumn{2}{c}{\textit{Sym.}} & \textit{Avg.} \\
    \cmidrule{3-6}\cmidrule{8-10}
    & & 50\% & 90\% & 40\% && & 50\%  & 90\%  \\
    \shline
    \ref{bl_1} & Baseline  & 70.0 & 20.6 &77.5 &    56.1  &&    47.3 &  6.8 &   27.1    \\
    \ref{bl_2} & Loss~\cite{arazo2019unsupervised,li2020dividemix} & 92.5 & 75.9 & 73.2 &  80.6  && 71.2 & 16.0 &  43.6   \\
    & $k$-NN~\cite{ortego2021multi} & 92.9 &     79.7 &  91.3 &  88.0  && 70.3 &     39.8 &  55.1   \\
            & OOD~(\bd{ours}) & 93.1 & 82.1 & 92.0 & 89.1  && 70.7 & 45.9 & 58.3  \\
    \ref{bl_3} & \emph{Ensem.}~\cite{liu2020early}  & 91.3 & 72.7 & 89.8 & 84.6 && 68.2 & 36.9 & 52.6 \\
            & $\mathcal{L}_{\mathrm{cross}}$~(\bd{ours})  & 93.9  & 89.4  & 92.6 & 92.0 && 73.3  & 54.5 & 63.9   \\
    \ref{bl_4} & w/o $\mathcal{L}_{\mathrm{reg}}$   & 92.0 & 34.5 & 90.3 & 72.3  && 68.5 & 24.3 & 46.4   \\
    \ref{bl_5} & w/o $\mathcal{L}_{\mathrm{align}}$   & 91.8 & 84.6 & 89.7 & 88.7 && 69.4 & 48.4 & 58.9   \\
    \ref{bl_6} & MoCo     & 94.4 &  90.7 &  93.1 &  92.7  &&    74.0 &  57.3 & 65.6   \\
    \shline
    \end{tabular}
    }
    \caption{
    Ablation results of different components in \methodname.}\label{tab:ablation_study}
   
\end{table}

\item\label{bl_4} \textbf{Without $\mathcal{L}_{\mathrm{reg}}$.}\quad
We remove $\mathcal{L}_{\mathrm{reg}}$ and the results indicate that it plays an important role, especially on extremely high label noise. Removing each term in $\mathcal{L}_{\mathrm{reg}}$ obtains similar results. We argue that $\mathcal{L}_{\mathrm{reg}}$ works in two aspects: 1) it can avoid the model collapse which outputs single classes, and 2) it can encourage the model to have high confidence for the predictions, which has shown its effectiveness for unlabeled data in semi-supervised learning.
\item\label{bl_5} \textbf{Without $\mathcal{L}_{\mathrm{align}}$.}\quad 
We remove $\mathcal{L}_{\mathrm{align}}$ and the performance has decreased, as expected, but is still more promising than other baselines. $\mathcal{L}_{\mathrm{align}}$ has leveraged mixup augmentation to regularize both classification and representation learning. Appendix~\ref{sec:knn} shows the evaluation of $k$-NN classification, demonstrating that $\mathcal{L}_{\mathrm{align}}$ can also greatly improve the learned representations.
\item\label{bl_6} \textbf{Contrastive Framework.}\quad
We implement \methodname into another contrastive framework for representation learning, \ie, MoCo~\cite{he2020momentum}. Based on the MoCo framework, our method has achieved more improvements in various experiments, which benefits from a large number of negative examples in the memory queue and a moving-averaged encoder~(we set the queue size and the factor of moving-average to 4,096 and 0.99, respectively).

\end{enumerate}

\begin{figure}
    \centering
    \includegraphics[width=0.75\linewidth]{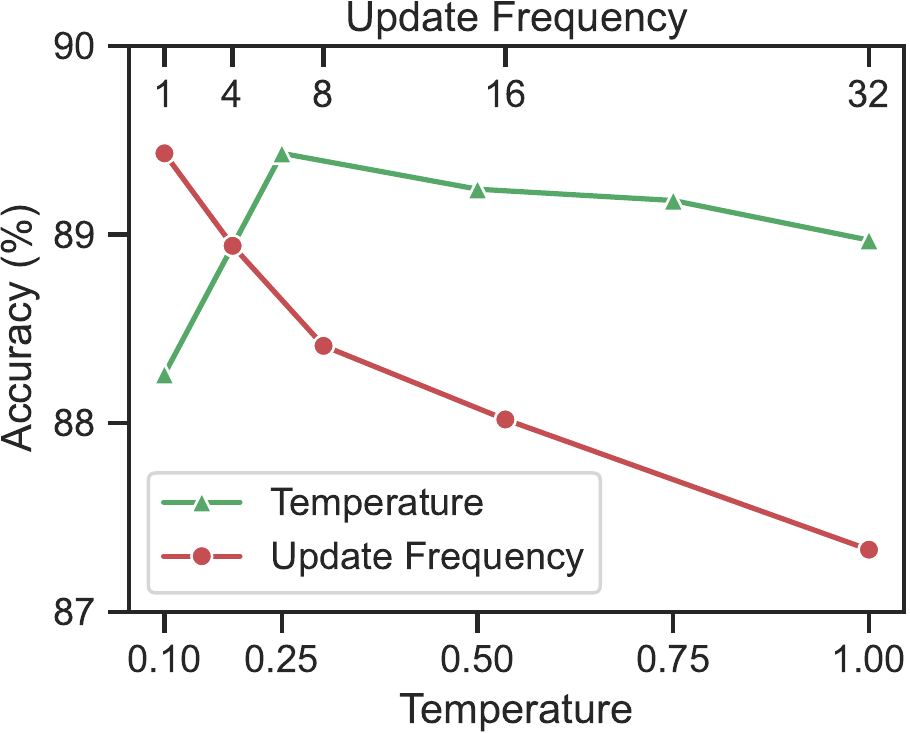}
    \caption{
        Ablation results for hyperparameters.
    }
    \label{fig:ablation_hyperparameters}%
\end{figure}

\paragraph{Hyperparameters.}
We evaluate the most essential hyperparameters  to our design, including the temperature $\tau$ for contrastive loss and update frequency for \methodname on CIFAR-10 with 90\% symmetric noise. Here, the update frequency denotes how may epochs that we update the parameters of \methodname, $\{(\vct{\mu}_k, \sigma_k)\}_{k=1}^K$ and $\{w_i\}_{i=1}^N$.
Fig.~\ref{fig:ablation_hyperparameters} shows that our method is robust to different choices of hyperparameters.
Even though \methodname updates for every 32 epochs, our method has still performed well, which indicates that the computational cost can be significantly reduced.

\subsection{Results on real-world datasets}

\paragraph{Datasets and training details.} 
We validate our method on two real-word noisy datasets: WebVision~\cite{li2017webvision} and Clothing1M~\cite{xiao2015learning}. Webvision contains millions of noisily-labeled images collected from the web using the keywords of ImageNet ILSVRC12~\cite{deng2009imagenet}. Following the convension~\cite{ortego2021multi,li2020dividemix,li2021learning,liu2020early}, we conducted experiments on the first 50 classes of the Google image subset, termed WebVision~(mini) and evaluated on both WebVision and ImageNet validation set. Clothing1M consists of 14 kinds of images
collected from online shopping websites. Only the noisy training set is used in our experiments.
We used a batch size of 256 on 4 GPUs, and trained a ResNet-50 for 40 epochs~(without warm-up) on Clothing1M and a ResNet-18 for 130 epochs~(warm-up 10 epochs) on WebVision, respectively. Following~\cite{li2020dividemix,li2021learning,liu2020early}, for Clothing1M, the encoder is initialized with ImageNet pre-trained weights, the initial learning rate is 0.01, and 256 mini-batches are sampled as one epoch. Other hyper-parameters are kept to be the same without further tuning.

\paragraph{Quantitative results.} 
Tables~\ref{tab:webvision} and~\ref{tab:clothing1m} present the results on WebVision and Clothing1M datasets. Our method outperforms state-of-the-art methods on both datasets, demonstrating its superior performance in handling real-world noisy datasets. We note that after checking the Clothing1M dataset, there are still lots of mislabeled images in the testing set. Therefore, the results on Clothing1M may not be such reliable as other datasets to evaluate the true performance of different methods.

\begin{table}
        \begin{tabular}{r|cc|cc}
               & \multicolumn{2}{c|}{\bd{WebVision}} & \multicolumn{2}{c}{\bd{ILSVRC12}} \\
               & top1 & top5 & top1 & top5 \\
            \shline
            Forward~\cite{patrini2017making}   & 61.1 & 82.6 & 57.3 & 82.3 \\
            D2L~\cite{ma2018dimensionality}   & 62.6 & 84.0 & 57.8 & 81.3 \\
            Iterative-CV~\cite{chen2019understanding}   & 65.2 & 85.3 & 61.6 & 84.9 \\
            Decoupling~\cite{malach2017decoupling}   & 62.5 & 84.7 & 58.2 & 82.2 \\
            MentorNet~\cite{jiang2018mentornet}   & 63.0 & 81.4 & 57.8 & 79.9 \\
            Co-teaching~\cite{han2018co}   & 63.5 & 85.2 & 61.4 & 84.7 \\
            ELR~\cite{liu2020early}   & 76.2 & 91.2 & 68.7 & 87.8 \\
            DivideMix~\cite{li2020dividemix}   & 77.3 &  91.6 & 75.2 & 90.8 \\
            RRL~\cite{li2021learning}   & 76.3 &  91.5 & 73.3 & 91.2 \\
            NGC~\cite{li2020mopro}   & \bd{79.1} & 91.8 & 74.4 & 91.0 \\
            MOIT~\cite{ortego2021multi}   & 77.9 & 91.9 & 73.8 &91.7 \\
            \hline
            \methodname (\bd{ours}) & \bd{79.1} & \bd{92.3} & \bd{75.4} & \bd{92.4} \\
            \end{tabular}
            \caption{Results on WebVision (mini).}\label{tab:webvision}
\end{table}
\begin{table}
    \begin{tabular}{rc}
        Method        & Acc~(\%) \\
        \shline
        Cross-Entropy  & 69.2      \\
        Label Correction~\cite{arazo2019unsupervised}      & 71.0      \\
        Joint-Opt~\cite{tanaka2018joint}   & 72.2      \\
        ELR~\cite{liu2020early}           & 72.8      \\
        SL~\cite{wang2019symmetric}  & 74.4      \\
        DivideMix~\cite{li2020dividemix}  & 74.4      \\
        MentorMix~\cite{jiang2020beyond}  & 74.3     \\
        RRL~\cite{li2021learning}  & \bd{74.8}     \\
        \hline
        \methodname (\bd{ours})                         & \textbf{74.8}  \\
    \end{tabular}
    \caption{Results on Clothing1M.
    }\label{tab:clothing1m}
\end{table}
% !tex root=./main.tex
\section{Conclusion}
In this paper, we introduced \methodname, a novel twin contrastive learning model for learning from noisy labels. By connecting the \emph{label-free} latent variables and \emph{label-noisy} annotations, \methodname can effectively detect the label noise and accurately estimate the true labels. 
Extensive experiments on both simulated and real-world datasets have demonstrated the superior performance of \methodname than existing state-of-the-art methods. 
In particular, \methodname achieves 7.5\% performance improvement under extremely 90\% noise ratio.
In the future, we will improve \methodname with semantic information for low noise ratios and explore dynamically updating the GMM.

\clearpage
%%%%%%%%% REFERENCES
{\small
\bibliographystyle{ieee_fullname}
% \bibliography{ref}

}
%!tex root=main.tex
\clearpage
% \onecolumn
\appendix
\renewcommand \thepart{}
\renewcommand \partname{}
% \part{\hfill \textsc{Appendix} \hfill}
\addcontentsline{toc}{section}{Appendix}
\numberwithin{equation}{section}
\setcounter{figure}{0}
\setcounter{table}{0}
 \renewcommand\thetable{\thesection\arabic{table}}

\section{$k$-NN evaluation}
\label{sec:knn}

We perform the $k$-NN classification over the learned representations with $k=200$.
For comparisons, we removed all proposed components and reported the performance on the representations learned in a pure unsupervised manner. The clean labels are involved in testing but excluded in the training phase.
The results are shown in Tables~\ref{tab:results_on_knn_cifar10} and~\ref{tab:results_on_knn_cifar100}.

The representations learned by our method have consistently outperformed the unsupervised learning, regardless of label noise with different ratios. These results indicate that our method has maintained meaningful representations better than the pure unsupervised learning model.

\begin{table}[ht]
\centering
\caption{
    $k$-NN evaluation on the learned representations of \methodname and unsupervised baseline on CIFAR-10.
}
\label{tab:results_on_knn_cifar10}
\scalebox{0.9}{
\begin{tabular}{rrrrrrr}
& \multicolumn{6}{c}{\bd{CIFAR-10}} \\
\multirow{2}{*}{Noise type/rate} & \multicolumn{4}{c}{\textit{Sym.}} & \textit{Asym.} & \textit{Avg.} \\ 
\cmidrule{2-6}
& 20\% & 50\% & 80\% & 90\% & 40\% & \\
\shline
$k$-NN~(\bd{ours}) & 94.9  & 94.0  & 92.2  & 90.6  & 92.8 & \bd{92.9} \\
$k$-NN~(\bd{unsup.}) & \multicolumn{5}{c}{---} & 86.4 \\
\end{tabular}
}
\end{table}

\begin{table}[ht]
    \centering
    \caption{
        $k$-NN evaluation on the learned representations of \methodname and unsupervised baseline on CIFAR-100.
    }
    \label{tab:results_on_knn_cifar100}
    \begin{tabular}{rrrrrr}
    
    & \multicolumn{5}{c}{\bd{CIFAR-100}} \\
    \multirow{2}{*}{Noise type/rate} & \multicolumn{4}{c}{\textit{Sym.}} & \textit{Avg.} \\ 
    \cmidrule{2-6}
    & 20\%  & 50\%  & 80\%  & 90\% & \\
    \shline
    $k$-NN~(\bd{ours}) & 76.7 & 72.6  & 67.3  & 64.1  & \bd{70.2} \\
    $k$-NN~(\bd{unsup.}) & \multicolumn{4}{c}{---} & 53.8\\
    \end{tabular}
\end{table}

\section{Asymmetric Label Noise}
\label{sec:asym_noise}

Table~\ref{tab:results_on_cifar_asymmetric} shows the results of TCL and TCL+ for CIFAR-10/100 under different asymmetric ratios following~\cite{li2022selective}, where our method has consistently outperformed the competitors.

We note that, unlike \emph{symmetric} label noise, the classes with above 50\% \emph{asymmetric} label noise cannot be distinguished, which makes 40\% becomes the most extreme scenario. In addition, we found that the asymmetric label noise would make the dataset imbalance, where the assumption of uniform distribution in Sec.~\ref{sec:model_data} does not hold.

Here, we employ the class imbalance ratio $r=\max(\{N_z\}_{z=1}^K)/\min(\{N_z\}_{z=1}^K)$ used in long-tailed learning to measure whether the label distribution is uniform, where $K$ and $N_z$ are the numbers of classes and samples in $z$-th class, respectively. 
The lower $r$ is, the more uniform the distribution becomes. 
For CIFAR-10 under the extreme high asymmetric label noise (\ie 40\%), $r=2.40$; that is, the asymmetric label noise makes the dataset non-uniform.
However, TCL can still achieve pleasing performance on non-uniform datasets, which suggests that TCL can effectively detect those mislabeled samples to form a uniform distribution.
Specifically, for those clean samples~(clean probability $w_i>0.5$), $r=1.37$, which is much more balanced over noisy labels.

\begin{table}[h]
    \centering
    \scalebox{0.8}{
    \begin{tabular}{rrrrrrrrr}
    & \multicolumn{3}{c}{\bd{CIFAR-10}} & & \multicolumn{4}{c}{\bd{CIFAR-100}} \\
    \cmidrule{2-4} \cmidrule{6-9}
    & 10\% & 20\% & 30\% & &10\% & 20\% & 30\% & 40\% \\
    \shline
    DivideMix~[20] & 93.8 & 93.2 & 92.5 & & 69.5 & 69.2 & 68.3 & 51.0 \\
    ELR~[25] & 94.4 & 93.3 & 91.5 & &  75.8 & 74.8 & 73.6 & 70.0 \\
    Sel-CL+~[23] & \underline{95.6} & \underline{95.2} & \underline{94.5} & & \underline{78.7} & \underline{77.5} & \underline{76.4} & \underline{74.2} \\
    \hline
    \methodname (\bd{ours}) & 95.1 & 94.7 & 94.4 & & 78.2 & 76.8 & 75.5 & 73.1 \\
    TCL+ (\bd{ours}) & \bd{95.9} & \bd{95.3} & \bd{94.8} & & \bd{79.0} & \bd{78.0} & \bd{76.9} & \bd{74.4} \\
    \shline
    \end{tabular}
    }
    \caption{
       Comparisons with SOTAs under \emph{asymmetric} label noise.
    }\label{tab:results_on_cifar_asymmetric}
 \end{table}
\end{document}